\documentclass[letterpaper, 10 pt, conference]{ieeeconf}  

\IEEEoverridecommandlockouts                              

\usepackage{amsmath,amssymb}
\usepackage{array}
\usepackage{cite}
\usepackage{amsfonts}
\usepackage{graphicx} 
\usepackage{epsfig} 
\usepackage{epstopdf}
\usepackage{epic}
\usepackage[percent]{overpic}
\usepackage{float}
\usepackage{algpseudocode}
\usepackage{algorithm}
\usepackage{footnote}
\usepackage{balance}
\usepackage{mathrsfs}
\usepackage{mathtools}
\usepackage{times} 
\usepackage{makeidx}
\usepackage{float}
\usepackage{url}
\usepackage[hidelinks]{hyperref}
\usepackage{flexisym}
\usepackage{chngcntr}
\usepackage{array}
\usepackage{subfigure}
\usepackage{color}
\usepackage[pdftex,dvipsnames]{xcolor}
\usepackage{lipsum} 
\usepackage{ulem}
\usepackage{xargs}
\usepackage{setspace}
\usepackage{cleveref}
\usepackage{multirow}
\usepackage{eurosym,rotating,booktabs}
\usepackage[flushleft]{threeparttable}
\usepackage{setspace}
\usepackage{textcomp}
\usepackage{footnote}
\usepackage{tikz}
\usepackage{xcolor,colortbl}
\usepackage{siunitx}
\usepackage{nccmath}

\newcommand{\simiid}[1]{\mathbin{\overset{#1}{{\tiny \textrm{\kern\z}} \sim}}}%

\algnewcommand\algorithmicforeach{\textbf{for each}}
\algdef{S}[FOR]{ForEach}[1]{\algorithmicforeach\ #1\ \algorithmicdo}
\algdef{SE}[FOR]{NoDoFor}{EndFor}[1]{\algorithmicfor\ #1}{\algorithmicend\ \algorithmicfor}
\algdef{SE}[IF]{NoThenIf}{EndIf}[1]{\algorithmicif\ #1}{\algorithmicend\ \algorithmicif}%
\algdef{SE}[WHILE]{NoDoWhile}{EndWhile}[1]{\algorithmicwhile\ #1}{\algorithmicend\ \algorithmicwhile}%

\newtheorem{problem}{Problem}

\DeclareSymbolFont{symbolsC}{U}{txsyc}{m}{n}
\DeclareMathSymbol{\notniFromTxfonts}{\mathrel}{symbolsC}{61}

\usepackage{ulem}
\usepackage{xargs}
\usepackage{cleveref}
\let\llncssubparagraph\subparagraph
\let\subparagraph\paragraph
\usepackage[compact]{titlesec}
\let\subparagraph\llncssubparagraph

\titlespacing*{\section}{0pt}{*1}{*1}
\titlespacing*{\subsection}{0pt}{*.5}{*1}

\def\BibTeX{{\rm B\kern-.05em{\sc i\kern-.025em b}\kern-.08em
    T \kern-.1667em\lower.7ex\hbox{E}\kern-.125emX}}

\usepackage{booktabs}
\usepackage{multirow}
\usepackage{array}
\usepackage[T1]{fontenc}
\usepackage[utf8]{inputenc} 
\usepackage{tabularx,ragged2e,booktabs}
\newcolumntype{C}[1]{>{\Centering}m{#1}}

\usepackage{makecell}
\usepackage{amsmath}
\usepackage{amsfonts}
\usepackage[inkscapelatex=false]{svg}

\DeclareMathOperator*{\argmin}{arg\,min}

\title{
Multi-Agent Informative Sampling 
using Distributed 
Gaussian Processes
}

\author{Sanket A. Salunkhe \and George P. Kontoudis
\thanks{The authors are with the Colorado School of Mines, Golden, CO 80401, USA. {\tt\footnotesize email: \{sanket\_salunkhe, george.kontoudis\}@mines.edu}.}%
}

\begin{document}

\maketitle
\thispagestyle{empty}
\pagestyle{empty}

\begin{abstract}

Multi-agent informative sampling 
aims to coordinate a team of agents to efficiently estimate latent fields. 
Gaussian Processes (GPs) provide a principled way for 
surrogate modeling due to their prediction and uncertainty quantification capabilities, but suffer from high 
complexity and 
scalability. 
In this paper, we propose a Multi-Agent Informative Sampling framework that integrates distributed GP learning, coordinated informative sampling, and trajectory planning in an 
iterative {learning-planning} loop.
Each agent constructs a 
compact pseudo-dataset that enables 
team-level probabilistic belief sharing via distributed consensus-based GP training. 
Moreover, we introduce a distributed optimization of Cohn's utility 
to coordinate multi-agent informative sampling. 
We evaluate our framework using real-world NASA SRTM elevation datasets across varying fleet sizes, and compare against coverage planners and GP-based sampling baselines.
The results 
demonstrate improved sample efficiency, faster error reduction, 
and better uncertainty calibration,  
highlighting the benefits of coupling distributed learning with multi-agent planning. 

\end{abstract}



\section{Introduction}

Cooperative multi-agent systems are essential for 
autonomous exploration~\cite{10876031} 
and long-duration environmental monitoring~\cite{dunbabin2012robots} missions. 
Distributing sensing and decision-making across multiple agents improves fault tolerance and scales to large environments beyond single-agent capabilities.
Most of the prior exploration work has focused on \textit{geometric coverage objectives} such as frontier-based exploration~\cite{mahdoui2018cooperative} and coverage path-planning strategies~\cite{Lin2021Collaborative}, which 
achieve complete or near-complete coverage. While coverage-based approaches are effective for mapping, 
they 
can be sample-inefficient. 
In such applications, \textit{Multi-agent Informative Sampling} explicitly optimizes measurement locations to maximize information gained about the latent phenomenon per unit cost, often outperforming coverage-based exploration under limited sensing and operational budgets~\cite{Cao2013Multi-robot}.

Informative sampling reformulates exploration as an information‑maximization problem, where agents select feasible sampling waypoint sequences that maximize information gain about an unknown environment~\cite{binney2010informative}.
The 
sampling locations are chosen using 
acquisition functions based on information-theoretic criteria such as expected entropy reduction, Fisher information, or integrated posterior variance~\cite{hollinger2014sampling,kontoudis2023adaptive}.
Gaussian Processes (GPs) have been used as surrogate models for informative sampling, because they provide a nonparametric prior with closed-form posterior predictions 
and calibrated uncertainty through the posterior predictive covariance~\cite{rasmussen2006gaussian}. 
Information metrics such as mutual information, entropy, and variance reduction can be computed directly from the predictive covariance structure~\cite{marchant2012bayesian}. 


Recent works in Multi-agent Informative Sampling for GP-based exploration have advanced scalable GP approximations, motion-constrained planning, and decentralized coordination under limited communication~\cite{jakkala2024multi,newaz2023decentralized,
le2021efficient}.
Notably,~\cite{jakkala2024multi} introduced a sparse GP-based informative sampling framework with continuous gradient-based path optimization under routing and sensing constraints, providing a strong baseline method for 
multi-agent sampling and planning.
A complementary line of work 
links sampling and routing decisions to reconstruction accuracy or posterior uncertainty~\cite{dutta2022approximation,jakkala2026informative}. 
To address communication constraints, decentralized strategies often localize information exchange and enable asynchronous coordination. Newaz et al.~\cite{newaz2023decentralized} exploit spatial structure to limit transmission of low-novelty data, 
while asynchronous variants 
combine Bayesian optimization, distributed control, and online hyperparameter learning to optimize resources~\cite{
le2021efficient}. 
Beyond target selection,~\cite{kailas2025integrating} integrates adaptive sampling with informative routing by selecting subgoals from mixtures of local GPs. 

Despite these advances, 
existing approaches rarely maintain a globally consistent probabilistic belief 
while 
coordinating 
across agents.
Consequently, agents either plan from partially consistent local beliefs or exchange large amounts of data, which creates scalability bottlenecks.  
To address this gap, we propose a unified Multi-agent Informative Sampling framework 
that couples distributed GP learning via compact pseudo-dataset exchange, 
coordinated informative 
sample selection, and trajectory planning.

\textit{Contribution:} 
First, we formulate a 
Multi-agent Informative Sampling (MIS) method that integrates distributed GP modeling, 
coordinated informative sampling, and trajectory planning in a single learning-planning iterative loop. 
Second, we develop a distributed optimization of Cohn’s utility to enable coordinated sampling across agents. 

\section{Problem Formulation}\label{sec:problem_formulation}


Consider a team of $M$ agents operating in a bounded space  $\mathcal{{X}}  \subset \mathbb{R}^D$, characterized by an unknown latent spatial field  $f: \mathcal{{X}} \to \mathbb{R}$. 
The objective of multi-agent informative sampling is to collectively sample locations that maximize the expected informative gain. The latent field $f$ is modeled as a continuous function over the input domain $\mathcal{{X}}$~\cite{hitz2017adaptive}. Gaussian Processes (GPs) provide a data-efficient method for estimating latent fields 
while simultaneously quantifying the predictive uncertainty. However, exact GP inference scales cubically $\mathcal{O}(n^3)$ with 
dataset size~$n$~\cite{rasmussen2006gaussian,gramacy2020surrogates}. This challenge is amplified in multi-agent systems where onboard computational resources are limited. 

\subsection{Gaussian Processes}\label{sec:gp}
Let us model the observations as $y(\boldsymbol{x}) = f(\boldsymbol{x}) + \epsilon$, where $\boldsymbol{x} \in \mathbb{R}^D$ is the input location of dimension $D$, $y(\boldsymbol{x}) \in \mathbb{R}$ the corresponding noisy scalar measurement, and $\epsilon \sim \mathcal{N}(0, \sigma_{\epsilon}^2)$ represents zero-mean Gaussian measurement noise with variance $\sigma_{\epsilon}^2$.
The unknown latent function is modeled as a zero-mean GP, $f(\boldsymbol{x}) \sim GP(0, k_{\boldsymbol{\theta}}(\boldsymbol{{x}},\boldsymbol{{x}}'))$, where 
$k_{\boldsymbol{\theta}}(\boldsymbol{{x}},\boldsymbol{{x}}')$ is the covariance parameterized by hyperparameters $\boldsymbol{\theta}$.
Given a dataset $\mathcal{D}=\{ \boldsymbol{X} ,\boldsymbol{y} \}$, where $\boldsymbol{X} = \left\{ \boldsymbol{x}_i \right\}_{i=1}^n \in \mathbb{R}^{n\times D}$ are the input locations 
and $\boldsymbol{y} = \left\{ {y}_i \right\}_{i=1}^n \in \mathbb{R}^{n}$ the 
noisy observations (or outputs), the hyperparameters $\boldsymbol{\theta}$ are trained by minimizing the negative marginal log-likelihood  function $\mathcal{L}(\boldsymbol{\theta})$ up to additive constants, 
\begin{align} \label{eq:nll_min_problem}
    \hat{\boldsymbol{\theta}} = \arg\min_{\boldsymbol{\theta}} \{\mathcal{L}(\boldsymbol{\theta}) \}=\arg\min_{\boldsymbol{\theta}} 
    \{ \boldsymbol{y}^{\intercal} \boldsymbol{C}_{\theta}^{-1} \boldsymbol{y} 
        +  \log |\boldsymbol{C}_{\theta}| \}, 
\end{align}
where $\boldsymbol{C}_{\theta} = \boldsymbol{K}+\sigma_{\epsilon}^2I_n \in \mathbb{R}^{n \times n}$ is the covariance matrix, and $\boldsymbol{K} = k_{\boldsymbol{\theta}}(\boldsymbol{X},\boldsymbol{X}) \in \mathbb{R}^{n \times n} $ the kernel matrix. Once we obtain an estimate of the hyperparameters~$\hat{\boldsymbol{\theta}}$, then the mean and variance of 
the posterior 
at a test input $\boldsymbol{x}_*$ becomes, 
\begin{align} 
    \mu(\boldsymbol{x}_*) &= k_*^{\intercal} \boldsymbol{C}_{\theta}^{-1}\boldsymbol{y} \label{eq:gp_pred_mean} \\ 
    \sigma^2(\boldsymbol{x}_*) &= k_{**} - k_*^{\intercal} \boldsymbol{C}_{\theta}^{-1}k_* ,\label{eq:gp_pred_variance}
\end{align} \label{eq:gp_pred_eq}
where $k_* =  k_{\boldsymbol{\theta}}(\boldsymbol{X},\boldsymbol{x}_*) \in \mathbb{R}^{n}$ and $k_{**}  = k_{\boldsymbol{\theta}}(\boldsymbol{x}_*,\boldsymbol{x}_*) \in \mathbb{R}$. 


\subsection{Distributed Gaussian Process}
In multi-agent systems, measurements are typically partitioned across agents. 
This partitioning renders global GP training on the aggregated dataset $\mathcal{D}=\bigcup_{i=1}^{M}\mathcal{D}_i$ computationally and communication-wise prohibitive. \textit{Distributed GP} methods instead 
aim to collectively estimate a global hyperparameter vector~$\hat{\boldsymbol{\theta}}$ while maintaining a shared team-level belief through local computations and limited communication. By distributing the dataset across $M$ agents, the per-agent computational cost of GP training is reduced from~$\mathcal{O}(n^3)$ to approximately $\mathcal{O}((n/M)^3)$. The factorized GP training (factGP)~\cite{deisenroth2015distributed} approximates the global objective~$\mathcal{L}(\boldsymbol{\theta})$ as a sum of local objectives~$\mathcal{L} \approx \sum_{i=1}^{M} \mathcal{L}_i$, a common approach in distributed optimization~\cite{yang2019survey}. 
Each agent $i$ first computes a local hyperparameter vector~$\boldsymbol{\theta}_i$ by performing GP training~\eqref{eq:nll_min_problem} on their local dataset $\mathcal{D}_i$. 

To mitigate these limitations, 
the generalized approximate proximal GP (gapxGP)~\cite{kontoudis2023decentralized,kontoudis2024scalable} and proximal inexact pseudo GP (pxpGP)~\cite{salunkhe2026federated}, 
augment local datasets~$\mathcal{D}_i$ using shared data summaries, either by 
randomly sampling local datasets~$\mathcal{D}_{+i}$ or by optimizing for pseudo-datasets~$\mathcal{D}_{+i}^{\star}$. Consensus over hyperparameters is enforced using the alternating direction method of multipliers (ADMM) 
\cite{boyd2011admm, shi2014linear}. 
The distributed optimization problem yields, 
\begin{align}\label{eq:admm_nll_min_problem}
    \hat{\boldsymbol{\theta}} =& \arg\min_{\boldsymbol{\theta}} \sum_{i=1}^{M} 
    \boldsymbol{y}_i^{\intercal} \boldsymbol{C}_{\boldsymbol{\theta}_i}^{-1} \boldsymbol{y}_i 
        +  \log |\boldsymbol{C}_{\boldsymbol{\theta}_i}|   \\ \nonumber
    \text{s.t.} \quad & \boldsymbol{\theta}_i = \boldsymbol{z}, \quad \forall i = 1, 2, \cdots, M,
\end{align}
where $\boldsymbol{z}$ is a shared slack variable. 
Next, each agent computes the local 
mean $\mu_i(\boldsymbol{x}_*)$~\eqref{eq:gp_pred_mean} and variance $\sigma_i^2(\boldsymbol{x}_*)$~\eqref{eq:gp_pred_variance}. 


\subsection{Informative Sampling}\label{sec:ipp}
Informative Sampling couples probabilistic exploration with planning by selecting measurement locations that maximize the expected information gain about an unknown latent field~\cite{hitz2017adaptive,binney2010informative}. At each informative sampling iteration $t$, the GP posterior, parameterized by the current hyperparameter estimate $\boldsymbol{\hat{\theta}}_t$ and conditioned on dataset $\mathcal{D}_t$, induces a predictive variance $\sigma_{t}^2(\boldsymbol{x})$ over the input domain $\mathcal{{X}}$. 
The informative sampling formulation employs an acquisition function 
$J$ 
and selects the next sampling location as,
\begin{align}\label{eq:ipp_opt}
    \boldsymbol{x}^{*}_{t+1} = \arg\max_{\boldsymbol{x} \in \mathcal{{X}}} J(\boldsymbol{\hat{\theta}}_t,\mathcal{D}_t;\boldsymbol{x}).
\end{align}
We employ the \textit{Cohn's acquisition function}~\cite{cohn1993neural,seo2000gaussian} that is defined as the expected reduction in variance 
over the entire input domain, 
\begin{align}\label{eq:cohns_acq_function}
    J_{\text{cohn}}(\boldsymbol{\hat{\theta}}_t, \mathcal{D}_t; \boldsymbol{x}) \coloneq \int_{\mathcal{X}} \Delta \sigma^2_{}(\boldsymbol{x}) \, \mathrm{d}x, 
\end{align}
where $\Delta \sigma^2_{}(\boldsymbol{x}) = \sigma^2_{t}(\boldsymbol{x}) - \tilde{\sigma}^2_{t+1}(\boldsymbol{x})$ denotes the variance reduction. Here, 
$\sigma^2_{t}(\boldsymbol{x})$ is the variance under the GP model conditioned on the current dataset $\mathcal{D}_t$, and $\tilde{\sigma}^2_{t+1}(\boldsymbol{x})$ is the approximated variance after augmenting $\mathcal{D}_t$ with the next potential location $\mathcal{D}_{t+1}=\mathcal{D}_t \cup (\boldsymbol{x}_{t+1},y_{t+1})$, while using the current hyperparameters $\boldsymbol{\hat{\theta}}_t$. 

The integral in~\eqref{eq:cohns_acq_function} is generally intractable over a continuous domain $\mathcal{X}$. Following~\cite{seo2000gaussian,kontoudis2023closed}, 
a quadrature approximation is used with a finite reference set $\boldsymbol{X}_{\text{ref}} \in \mathbb{R}^{N_{\text{ref}} \times D}$, where $N_{\text{ref}}$ denotes the number of reference points used to discretize $\mathcal{X}$. 
Thus, the optimization~\eqref{eq:ipp_opt} using \eqref{eq:cohns_acq_function} becomes,
\begin{align}\label{eq:cohns_optimization}
    \boldsymbol{x}^{*}_{t+1} \approx \arg&\min_{\boldsymbol{x} \in \mathcal{X}} \left\{ - \frac{1}{N_{\text{ref}}} \sum_{r=1}^{N_{\text{ref}}} \sigma^2_{t}(\boldsymbol{x}_r) - \tilde{\sigma}^2_{t+1}(\boldsymbol{x},\boldsymbol{x}_r) \right\}.
\end{align}
The predictive variances 
are given by,
\begin{align}
    \sigma^2_{t}(\boldsymbol{x}_r) &= k_r - \boldsymbol{k}_{n,r}^{\intercal} \boldsymbol{C}_n^{-1} \boldsymbol{k}_{n,r} \\
    \tilde{\sigma}^2_{t+1}(\boldsymbol{x},\boldsymbol{x}_r) &= k_r - \boldsymbol{k}_{n+1,r}^{\intercal} \boldsymbol{C}_{n+1}^{-1} \boldsymbol{k}_{n+1,r}
\end{align}
where
$k_r = k_{\boldsymbol{\hat{\theta}}}(\boldsymbol{x}_r, \boldsymbol{x}_r) \in \mathbb{R}$,
$\boldsymbol{k}_{n,r} = k_{\boldsymbol{\hat{\theta}}}(\boldsymbol{x}_r, \boldsymbol{X}_n) \in \mathbb{R}^n$,
$\boldsymbol{k}_{n+1,r} = k_{\boldsymbol{\hat{\theta}}}(\boldsymbol{x}_r, \boldsymbol{X}_{n+1}) \in \mathbb{R}^{n+1}$,
$\boldsymbol{C}_n = k_{\boldsymbol{\hat{\theta}}}(\boldsymbol{X}_{n}, \boldsymbol{X}_{n})+\sigma_{\epsilon}^2I_n \in \mathbb{R}^{n\times n}$, and
$\boldsymbol{C}_{n+1} = k_{\boldsymbol{\hat{\theta}}}(\boldsymbol{X}_{n+1}, \boldsymbol{X}_{n+1})+\sigma_{\epsilon}^2I_{n+1} \in \mathbb{R}^{(n+1)\times (n+1)}$
with $\boldsymbol{X}_{n+1} = \begin{bmatrix} \boldsymbol{X}_{n}^{\intercal} & \boldsymbol{x} \end{bmatrix}^{\intercal} \in \mathbb{R}^{(n+1)\times D}$ the augmented input set. 
Using the rank-one update of the GP variance reduction~\cite{seo2000gaussian}, 
\eqref{eq:cohns_optimization} can be written as,
    \begin{align}\label{eq:cohns_optimization_final}
        &\boldsymbol{x}_{t+1}^{*} = \arg\min_{\boldsymbol{x} \in \mathcal{X}}  \left\{ - \frac{1}{N_{\text{ref}}} \sum_{r=1}^{N_{\text{ref}}} \frac{(\boldsymbol{k}_{n,r}^{\intercal} \boldsymbol{C}_n^{-1} \boldsymbol{k}_{n+1} - k_{\boldsymbol{\hat{\theta}}}(\boldsymbol{x}, \boldsymbol{x}_r))^2}{k_{n+1,n+1} - \boldsymbol{k}_{n+1}^{\intercal} \boldsymbol{C}_n^{-1} \boldsymbol{k}_{n+1}} \right\},
    \end{align}
where $k_{n+1,n+1} = k_{\boldsymbol{\hat{\theta}}}(\boldsymbol{x}, \boldsymbol{x}) \in \mathbb{R}$,  $\boldsymbol{k}_{n+1} = k_{\boldsymbol{\hat{\theta}}}(\boldsymbol{x}, \boldsymbol{X}_n) \in \mathbb{R}^{n}$.


\begin{problem}\label{problem:mripp}
    Consider a team of $M$ agents operating in a bounded domain $\mathcal{X}\subset \mathbb{R}^D$ characterized by an unknown latent field $f:\mathcal{X}\rightarrow \mathbb{R}$. At iteration $t$, given the current local datasets $\mathcal{D}_t = \cup_{i=1}^M\mathcal{D}_{i,t}$ and the corresponding GP surrogate model parametrized by $\hat{\boldsymbol{\theta}}_t$, determine a coordinated set of input locations $\boldsymbol{S}_{t+1} = \{ \boldsymbol{x}_{1,t+1}^{*}, \cdots, \boldsymbol{x}_{M,t+1}^{*} \}\subset \mathcal{X}$ such that the joint measurements collected by the team maximize the expected uncertainty reduction over the entire domain $\mathcal{X}$.
\end{problem}

\section{Proposed Approach}\label{sec:proposed_approach}

We propose a Multi-Agent Informative Sampling framework that integrates:
(i) 
distributed GP modeling of an unknown field, and (ii) distributed optimization of Cohn’s acquisition function for coordinated sample selection.
At each sampling iteration, the agents update a globally consistent GP belief via distributed hyperparameter optimization and they cooperatively select a batch of informative sampling points using a distributed acquisition utility. 

\subsection{Distributed GP Training}\label{subsec:dist_gp}
To evaluate acquisition utilities under a globally consistent GP belief over $f$, we employ pxpGP~\cite{salunkhe2026federated}. Each agent constructs a compact pseudo-dataset~$\mathcal{D}_{i,t}^{*}$ from its local measurements and optimizes hyperparameters via a consensus-based, scaled proximal-inexact ADMM (pxADMM) with convergence guarantees~\cite{xie2019distributed}. We select the Separable Squared Exponential (SSE) kernel, 
\begin{align}\label{eq:sse_kernel}
    k(\boldsymbol{x} ,\boldsymbol{x}^{\prime}) = \sigma_f^2 \exp \left\{ -   \frac{1}{2} \sum_{d=1}^D\frac{\left({x}_d - {x}_d^{\prime}\right)^2 }{l_d^2}\right\},
\end{align}
with hyperparameters $\boldsymbol{\theta} = 
\left[ l_1, l_2, \cdots, l_D, \sigma_{f}, \sigma_{\epsilon} \right]^{\intercal} \in \mathbb{R}^{D+2}$, where $\sigma_f > 0$ and $l_d > 0$ denote the signal variance and the $d$-th lengthscale, respectively. 

During initialization, each agent $i$ fits a local sparse variational GP (SVGP)~\cite{titsias2009variational,mavridis2022sparse} 
to its local seed dataset~$\mathcal{D}_{i,0}^{}$ yielding a compact pseudo-summary~$\mathcal{D}_{i,0}^{*}$ of
$P$ pseudo-points.
This is achieved by 
minimizing the negative Evidence Lower Bound (ELBO) over the variational distribution $q(f_P)=\mathcal{N} \left( \boldsymbol{\mu}_P, \boldsymbol{A}_P \right)$ and inducing locations $\boldsymbol{X}_P$, 
\begin{align}\nonumber
    &q(f_P) = \min_{q(f_P), \boldsymbol{X}_P} - \mathfrak{L}_{\text{ELBO}}\\ \label{eq:min_elbo_divergence}
    &= \min_{q(f_P), \boldsymbol{X}_P} - \left( \mathbb{E}_{q(\boldsymbol{f})} \left[ \text{log } p(\boldsymbol{y|f})\right]  - \text{KL} \left( q(\boldsymbol{f}_P) \mid \mid p(\boldsymbol{f}_P) \right) \right), 
\end{align}
where $\text{KL}(\cdot \| \cdot)$ denotes the Kullback-Leibler divergence from the GP prior $p(\boldsymbol{f}_P)$ to $q(\boldsymbol{f}_P)$.
To prevent unconstrained inducing points from drifting outside data bounds or clustering in dense regions, we incorporate the boundary ($\mathfrak{L}_{b}$) and repulsive penalty ($\mathfrak{L}{r}$), 
\begin{align}\label{eq:elbo_final_obj}
    \boldsymbol{X}_P =\argmin_{q(f_P), \boldsymbol{X}_P} 
    - \mathfrak{L}_{\text{ELBO}} + \mathfrak{L}_{b} + \mathfrak{L}_{r}.
\end{align}

In the scaled pxADMM formulation, the augmented Lagrangian is locally linearized around the shifted variable, yielding the distributed GP hyperparameter updates. 
Each local model warm-starts using previous converged hyperparameters 
during initialization. 
Convergence is monitored using primal $\boldsymbol{r}_i^{(s+1)}$ and dual $\boldsymbol{s}_i^{(s+1)}$ residuals, employing adaptive penalty balancing $\rho_i$ and an Armijo backtracking line search for the Lipschitz parameter $L_i^{(s+1)}$~\cite
{salunkhe2026federated}.


\subsection{Proposed Distributed Cohn's Acquisition Function}\label{ssec:dist_cohn}
At each sampling iteration $t$, the distributed GP update provides a shared posterior uncertainty. 
Although the team reaches consensus on $\boldsymbol{\hat{\theta}}_t$, each agent conditions on its own augmented dataset $\mathcal{D}_{+i,t}$. Therefore, local objectives $J_{i,t+1}$ differ and require consensus over $\boldsymbol{x}$.
We leverage this team-level belief to extend the single-agent Cohn criterion~\eqref{eq:cohns_optimization_final} to the multi-agent setting. 
Sample selection is coupled, since the value of one sample depends on which samples have already been selected. Instead of a single optimization returning an $M$-batch point set, we construct the sample set~$\boldsymbol{S}_{t+1} = \{ \boldsymbol{x}_{m,t+1}^{*} \}_{m=1}^{M}$ \textit{sequentially}. 
Thus, we solve the following optimization problem $M$ times,
    \begin{align}
    \boldsymbol{x}_{m,t+1}^{*} &= \arg\min_{\boldsymbol{x} \in \mathcal{X}}\sum_{i=1}^{M}  -\int_{\mathcal{X}} \Delta \sigma^2_{i}(\boldsymbol{x}) \, \mathrm{d}x. \label{eq:dist_cohn_opt_integral}
\end{align}
We construct the sample set~$\boldsymbol{S}_{t+1}$ using a conditioned sequential greedy procedure. 
For example, at greedy step $m=3$, the next sample~$\boldsymbol{x}_{3,t+1}^{*}$ 
is selected by using~\eqref{eq:dist_cohn_opt_integral}, conditioned on all previously selected samples~$\{\boldsymbol{x}_{m,t+1}^{*}\}_{m=1}^2$. 
After each selection, all agents update their GP posterior {using a covariance-only update} to account for the newly chosen sample. 
This update mitigates redundant selections and prevents duplicate selections. 
Since the batch size matches the fleet size~$|\mathcal{V}|=M$, 
we obtain an one-to-one assignment between $M$~agents and $M$~sampling locations $\{\boldsymbol{x}_{m,t+1}^{*}\}_{m=1}^M$. 

Directly solving Cohn’s utility~\eqref{eq:dist_cohn_opt_integral} over the continuous domain is intractable and requires approximating the domain integral. 
Thus, we use a finite \textit{reference set}~$\boldsymbol{X}^{\text{ref}}_{t}$ to approximate the integral, and a \textit{candidate set}~$\boldsymbol{X}^{\text{cand}}_{t}$ to initialize optimization in multiple start locations. Each agent $i$ draws local samples~$\boldsymbol{X}^{\text{ref}}_{i,t}$ and~$\boldsymbol{X}^{\text{cand}}_{i,t}$ using the Latin Hypercube Sampling (LHS) method~\cite{gramacy2020surrogates} within its local region, while excluding previous augmented dataset~$\mathcal{D}_{+i,t}$ and enforcing a minimum separation. 
Thus, the distributed objective~\eqref{eq:dist_cohn_opt_integral} becomes,
\begin{align}\label{eq:dist_cohns_optimization}
    \boldsymbol{x}^{*}_{m,t+1} 
    =\arg\min_{\boldsymbol{x} \in \mathcal{X}} \left\{ \sum_{i=1}^{M} \left( - \frac{1}{N_{\text{ref}}} \sum_{r=1}^{N_{\text{ref}}} \Delta \sigma^2_{i}(\boldsymbol{x}, \boldsymbol{x}_r)
    \right) \right\},
\end{align}
where $\Delta \sigma^2_{i}(\boldsymbol{x},\boldsymbol{x}_r) = 
    \sigma^2_{i,t}(\boldsymbol{x}_r) - \tilde{\sigma}^2_{i,t+1}(\boldsymbol{x},\boldsymbol{x}_r)$ is the predicted variance reduction at reference location $\boldsymbol{x}_r \in \boldsymbol{X}^{\text{ref}}_{t}$. 
    Using the standard rank-one GP variance-reduction form,~\eqref{eq:dist_cohns_optimization} yields, 
\begin{align}
    \Delta \sigma^2_{i}(\boldsymbol{x},\boldsymbol{x}_r) &= 
     \frac{(\boldsymbol{k}_{n+1,r;t}^{\intercal} \boldsymbol{C}_{n;t}^{-1} \boldsymbol{k}_{n+1;t} - k_{\boldsymbol{\hat{\theta}}_t}(\boldsymbol{x}, \boldsymbol{x}_r))^2}{k_{n+1,n+1;t} - \boldsymbol{k}_{n+1;t}^{\intercal} \boldsymbol{C}_{n;t}^{-1} \boldsymbol{k}_{n+1;t}}, \nonumber
\end{align}
where $\boldsymbol{k}_{n+1,r;t} = k_{\boldsymbol{\hat{\theta}}_t}(\boldsymbol{x}_r, \boldsymbol{X}_{n+1}) \in \mathbb{R}^{n+1}$,
$\boldsymbol{C}_{n;t} = k_{\boldsymbol{\hat{\theta}}_t}(\boldsymbol{X}_{n}, \boldsymbol{X}_{n})+\sigma_{\epsilon}^2I_n \in \mathbb{R}^{n\times n}$,
$\boldsymbol{k}_{n+1;t} = k_{\boldsymbol{\hat{\theta}_t}}(\boldsymbol{x}, \boldsymbol{X}_n) \in \mathbb{R}^{n}$, and
$k_{n+1,n+1;t} = k_{\boldsymbol{\hat{\theta}}_t}(\boldsymbol{x}, \boldsymbol{x}) \in \mathbb{R}$.

To solve 
\eqref{eq:dist_cohns_optimization} in a distributed manner, we employ the scaled pxADMM 
(Section~\ref{subsec:dist_gp}). Each agent evaluates Cohn's criterion locally 
$J_{i,t+1}(\boldsymbol{x})$, and the team solves 
~\eqref{eq:dist_cohns_optimization}. 
The pxADMM linearizes an augmented Lagrangian of $J_{i,t+1}$ around a stationary point~$\boldsymbol{c}_{i} = \boldsymbol{e} - \boldsymbol{h}_{i}$ that yields,
\begin{align}\nonumber
        \mathscr{L} \left(  \boldsymbol{x}_{i} , \boldsymbol{e}, \boldsymbol{c}_{i} \right) &=  \sum_{i=1}^{M} {J}_{i,t+1}(\boldsymbol{c}_{i}) +  \nabla_{\boldsymbol{x}}^{\intercal} {J}_{i,t+1}(\boldsymbol{c}_{i})  \left( \boldsymbol{x}_{i} - \boldsymbol{c}_{i} \right)  \\ \label{eq:cohns_augmented_lagrangian_equation}
        &\quad \quad \quad \quad + \frac{L_{i;J} + \rho_{i;J}}{2} \left\|  \boldsymbol{x}_{i}  - \boldsymbol{c}_{i} \right\|^2,
\end{align}
with local Lipschitz parameter $L_{i;J} > 0$ and penalty parameter $\rho_{i;J}$. 
The resulting iterative synchronous updates are,
\begin{subequations}
\begin{align}
    \boldsymbol{x}_{i}^{(s+1)} &= \boldsymbol{c}_{i}^{(s)} - \frac{1}{L_{i;J}^{(s)} + \rho_{i;J}^{(s)}} \nabla_{\boldsymbol{x}} {J}_{i,t+1}\left(\boldsymbol{c}_{i}^{(s)}\right) \label{eq:cohns_x_update} \\
    \boldsymbol{e}^{(s+1)} &= \frac{1}{M} \sum_{i=1}^{M} \left( \boldsymbol{x}_i^{(s+1)} + \boldsymbol{h}_i^{(s)} \right) \label{eq:cohns_z_update} \\
    \boldsymbol{h}_i^{(s+1)} &= \boldsymbol{h}_i^{(s)} + \boldsymbol{x}_i^{(s+1)} - \boldsymbol{e}^{(s+1)}. \label{eq:cohns_u_update}
\end{align}
\label{eq:cohns_iterative_scheme}%
\end{subequations}

We initialize $\boldsymbol{x}_i^{(0)}$ from the candidate set~$\boldsymbol{X}^{\text{cand}}_{t}$ and clamp iterates to the workspace~$\mathcal{X}$ bounds. Convergence is monitored using the primal~$\boldsymbol{p}_i^{(s+1)} = \boldsymbol{x}_i^{(s+1)} - \boldsymbol{e}^{(s+1)}$ and dual residuals~$\boldsymbol{d}_i^{(s+1)} = \rho_{i;J} \left( \boldsymbol{e}^{(s+1)} - \boldsymbol{e}^{(s)}\right)$. 
We further employ residual balancing to update~$\rho_{i;J}$ and an Armijo-style backtracking rule for~$L_{i;J}^{(s+1)}$.
After selecting a batch of $M$ samples, i.e., $\boldsymbol{S}_{t+1}$, agents find an one-to-one assignment with minimum travel cost. We use a sequential greedy distance-based task assignment, based on the agents’ current poses $\boldsymbol{p}_{i,t}$, 
yielding 
a unique sample $\boldsymbol{g}_{i, t+1}$ to each agent. 

\subsection{Trajectory Planning}\label{ssec:traj-plan}
Given an assigned goal $\boldsymbol{g}_{i,t+1}$, agent $i$ plans a straight-line trajectory $\boldsymbol{\tau}_{i,t+1}$ parameterized by a trapezoidal velocity profile subject to velocity and acceleration limits. The trajectory module returns the trajectory duration $T_{i,t+1}$, path length $P_{i,t+1}$ and a continuous reference path~$\{\boldsymbol{x}^{\text{traj},\ell}_{i,t+1}\}_{\ell=1}^{T}$.
After executing locally planned trajectories
each agent updates its augmented dataset~$\mathcal{D}_{+i,t}$ with its samples~$(\boldsymbol{x}_{i,t+1}^{*}, y_{i,t+1}^{*})$. 
The agent team then retrains the shared GP via distributed pxADMM, 
warm-started from the previous round’s hyperparameters~$\boldsymbol{\hat{\theta}}_{i,t+1}^{(0)} = \boldsymbol{\hat{\theta}}_{i,t}$. The updated team-level posterior is employed in the next sampling iteration. 

\begin{algorithm}[t!]
\caption{Multi-agent Informative Sampling}\label{algo:mripp}
{\small\textbf{Input:} 
$\text{ADMM}(\rho_i, L_i 
)$, Seed-data $\mathcal{D}_{i,0}$, Robot Pose $\boldsymbol{p}_{i,t}$, $\textrm{ITER}$

\textbf{Output:} $\boldsymbol{\hat{\theta}}$, $\mathcal{D}_{+i,t+1}$
}
\begin{algorithmic}[1]
\small
\State $\mathcal{D}_{i,0}^{*} \gets \texttt{trainSparseGP}(\mathcal{D}_{i,0}^{})$ \Comment{Initialization}
\State Communicate $\mathcal{D}_{i,0}^{*}$ to all agents
\State $\mathcal{D}_{+i,0}^{} = \mathcal{D}_{i,0}^{} \cup \left\{\bigcup_{j=1}^M \mathcal{D}_{j,0}^{*}\right\}$

\For{$t=1, \ldots, \textrm{ITER}$}
    \State $\boldsymbol{S}_{t+1} \gets \emptyset$ 
    \ForAll{$i=1,\dots,M$} \Comment{Distributed GP Training}
        \State $\boldsymbol{C}_{\theta,i}^{-1}, \boldsymbol{\hat{\theta}}^{}_{i,t} \gets \texttt{pxpGP}(\mathcal{D}_{+i,t}^{})$
        \State $(\boldsymbol{X}^{\text{ref}}_{t}, \boldsymbol{X}^{\text{cand}}_{t}) \gets \texttt{condLHS}(\mathcal{D}_{+i,t})$
    \EndFor
    
    \ForAll{$m=1,\dots,M$}    \Comment{Distributed Cohn}
        \State $\boldsymbol{x}_{m,t+1}^{*} \gets \texttt{distCohn}(\boldsymbol{X}^{\text{ref}}_{t}, \boldsymbol{X}^{\text{cand}}_{t}, \boldsymbol{\hat{\theta}})$~\eqref{eq:dist_cohns_optimization}
        \State $\texttt{updateGPPosterior}(\boldsymbol{x}_{m,t+1}^{*})$
        \State $\boldsymbol{S}_{t+1} = \boldsymbol{S}_{t+1} \cup \{\boldsymbol{x}_{m,t+1}^{*}\}$
        \State $(\boldsymbol{X}^{\text{ref}}_{t}, \boldsymbol{X}^{\text{cand}}_{t}) \gets \texttt{condLHS}(\mathcal{D}_{+i,t},\boldsymbol{x}_{m,t+1}^{*})$
    \EndFor
    
    \State $\boldsymbol{x}_{i,t+1}^{*} \gets \texttt{greedyTaskAssignment}(\boldsymbol{S}_{t+1})$
    \State $\boldsymbol{\tau}_{i,t+1}, T_{i,t+1}
    \gets \texttt{generateTraj}
    (\boldsymbol{p}_{i,t}^{}, \boldsymbol{x}_{i,t+1}^{*} )$ 
        \State $y_{i,t+1}^{*} \gets \texttt{getMeasurement}(\boldsymbol{x}_{i,t+1}^{*})$
    \State $\mathcal{D}_{+i,t+1} = \mathcal{D}_{+i,t} 
    \cup \left\{\bigcup_{i=1}^M (\boldsymbol{x}_{i,t+1}^{*}, y_{i,t+1}^{*}) \right\}$
    \State $\boldsymbol{\hat{\theta}} = \boldsymbol{\hat{\theta}}_{i,t+1}$ 
\EndFor\\
\Return $\boldsymbol{\hat{\theta}}$, $\mathcal{D}_{+i,t+1}$
\end{algorithmic}
\end{algorithm}
\subsection{Algorithm Overview}\label{subsec:algo_overview}

Algorithm~\ref{algo:mripp} summarizes the proposed Multi-agent Informative Sampling framework. 
During initialization, each agent collects a small seed dataset~$\mathcal{D}_{i,0}^{}$, typically via a short local exploration pass. 
It then fits a local sparse variational GP 
and compresses the seed data into a compact pseudo-summary~$\mathcal{D}_{i,0}^{*}$.
To ensure a shared starting belief among all agents, each agent exchanges these pseudo-summaries with the team and forms a local augmented dataset~$\mathcal{D}_{+i,0}^{} = \mathcal{D}_{i,0} \cup \left\{\bigcup_{j=1}^M \mathcal{D}_{j,0}^{*}\right\}$, $j\neq i$. Using $\mathcal{D}_{+i,t}$, the team performs distributed GP training with pxpGP 
to obtain a global GP belief (Section~\ref{subsec:dist_gp}).
At each sampling iteration $t$, agents construct a local {reference set} $\boldsymbol{X}_{i,t}^{\text{ref}}$ 
and {candidate set} $\boldsymbol{X}_{i,t}^{\text{cand}}$ 
using 
LHS 
and then aggregate a global reference set $\boldsymbol{X}^{\text{ref}}_t = \bigcup_{i=1}^M \boldsymbol{X}_{i,t}^{\text{ref}}$ and a candidate set $\boldsymbol{X}^{\text{cand}}_t = \bigcup_{i=1}^M \boldsymbol{X}_{i,t}^{\text{cand}}$.
Using the current GP surrogate model $\boldsymbol{\hat{\theta}}$ and 
$(\boldsymbol{X}^{\text{ref}}_t, \boldsymbol{X}^{\text{cand}}_t)$, the team selects a set of next $M$ informative samples $\boldsymbol{S}_{t+1} = \{\boldsymbol{x}_{1,t+1}^{*}, \cdots, \boldsymbol{x}_{m,t+1}^{*}\}$ sequentially by repeatedly optimizing Cohn's acquisition function~\eqref{eq:dist_cohns_optimization} and updating the GP posterior after each selected sample. 
Each minimization is solved using 
\eqref{eq:cohns_iterative_scheme}. 
The global sample set~$\boldsymbol{S}_{t+1}$ is then assigned to agents one-to-one via a sequential greedy distance-based task allocation, 
yielding per-agent goal~$\boldsymbol{g}_{i,t+1}$.
Then, each agent plans and executes a feasible trajectory~$\boldsymbol{\tau}_{i,t+1}$ to its goal $\boldsymbol{x}_{i,t+1}^{*}$, collects measurement $\boldsymbol{y}_{i,t+1}^{*}$.
The local augmented dataset $\mathcal{D}_{+i,t}$ is then updated with the agent’s informative samples, 
resulting in~$\mathcal{D}_{+i,t+1} = \mathcal{D}_{+i,t} \cup \left\{\bigcup_{i=1}^{M} (\boldsymbol{x}_{i,t+1}^{*}, y_{i,t+1}^{*}) \right\}$.
This informative sampling loop repeats until a fixed number of iterations $\textrm{ITER}$. 

\tikzset{every picture/.style={line width=0.75pt}} 
\begin{figure*}[htbp]
\centering
\input{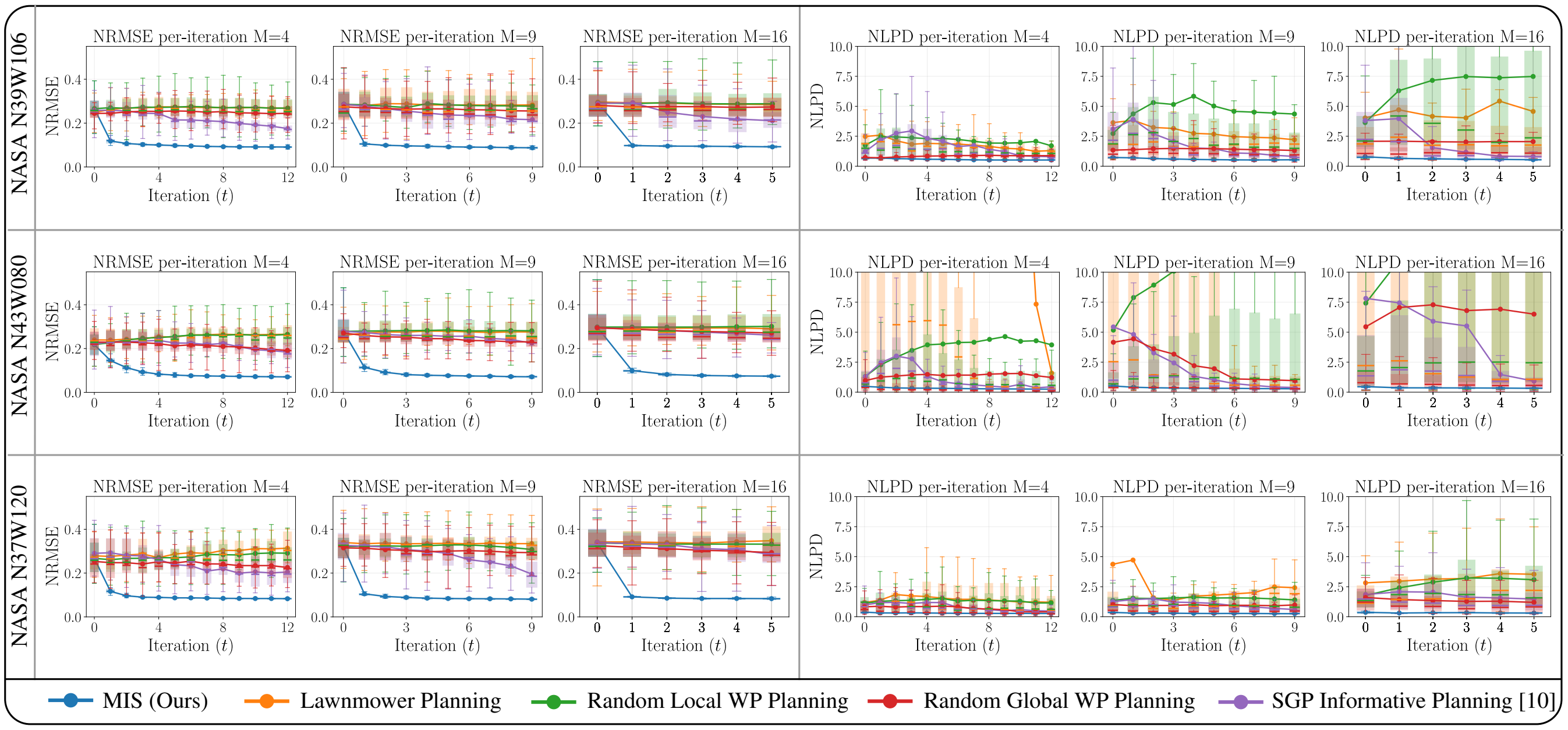}
\vspace{-20pt}
\caption{Comparison of prediction accuracy, 
  between the proposed Multi-agent Informative Sampling framework versus Lawnmower, Random Local Waypoints, Random Global Waypoints, and SGP-based IPP~\cite{jakkala2024multi} across fleet sizes $M \in \{4, 9, 16\}$ on three NASA SRTM datasets. 
  }
\label{fig:nrmse_nlpd_result}
\vspace{-10pt}
\end{figure*}
\vspace{-1pt}


\section{Experiments and Results}\label{sec:Experimental_Results}

We evaluate the proposed Multi-agent Informative Sampling (MIS) framework on {real-world terrain reconstruction tasks} using NASA Shuttle Radar Topography Mission (SRTM) elevation data across three tiles (N39W106, N37W120, and N43W080)~\cite{farr2007shuttle}. We compare the MIS 
against four baselines: coverage-based lawnmower
planning, local random waypoint planning, global random waypoint planning, and SGP-based multi-agent informative sampling~\cite{jakkala2024multi}. 
All methods use the same kernel~\eqref{eq:sse_kernel} and GP likelihood. The proposed MIS employs distributed pxpGP training, which improves computational efficiency and scalability at some cost in predictive accuracy and uncertainty quantification relative to standard GP training. Thus, the comparison is conservative, since MIS operates on an approximate 
GP surrogate 
model.

\begin{figure*}[htbp]
\centering
\input{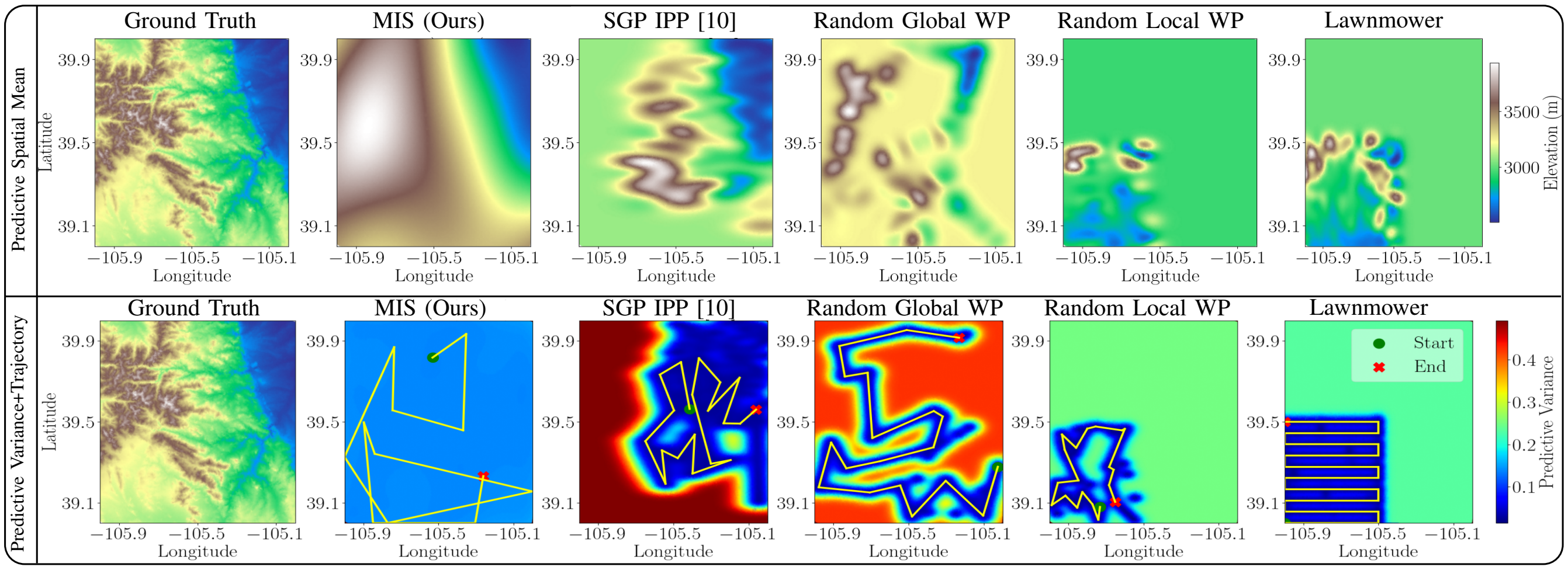}
\vspace{-20pt}
\caption{Ground-truth elevation map of 
  tile N39W106 and final individual agents GP posterior maps with trajectories for a fleet of $M=4$ agents.
  }
\label{fig:spatial_mean_traj}
\vspace{-10pt}
\end{figure*}

All experiments are implemented in Python using PyTorch and GPyTorch~\cite{gardner2018gpytorch}, and run on a workstation with an Intel Core i7-14700 CPU, 62 GB RAM, and an NVIDIA GeForce RTX 4080 GPU with 16 GB VRAM.
We evaluate performance using two standard GP regression metrics, (i) normalized root mean square error (NRMSE) and (ii) negative log predictive density (NLPD), reported at each planning iteration (Fig.~\ref{fig:nrmse_nlpd_result}). 
We also provide qualitative comparisons using per-agent predictive mean and variance maps (Fig.~\ref{fig:spatial_mean_traj}).

To evaluate scalability, we benchmark all methods with {
fully connected 
teams of size  $M \in \{4, 9, 16\}$. } 
Since larger teams collect data more rapidly per planning round, we scale the number of planning iterations 
accordingly, using $t=12, 9, 6$ iterations for $M=4, 9, 16$, 
where each planning iteration corresponds directly to selecting and measuring a single new sampling location. 
For all methods, we use the same initialization with a common seed dataset $\mathcal{D}_{i,0}$ containing a total of $n_0 = 10D = 20$ samples distributed across the team~\cite{gramacy2020surrogates}. 
All experiments are replicated over $10$ random seeds.
At each iteration, we evaluate the current model on a held-out test set of $n_{\text{test}} = 300$ samples and compute NRMSE and NLPD to track prediction accuracy and uncertainty calibration 
over time.

The pxADMM optimizer is initialized with $\rho^{(0)} = 1.0$ and $L_i^{(0)} = 2.0$ 
and a maximum of $s^{\text{max}} = 50$ ADMM iterations for both GP training 
and Cohn's optimization~\eqref{eq:dist_cohns_optimization}. For compact pseudo-summaries, we cap the pseudo-dataset size to $10\%$ of the corresponding local dataset for the seed data. 
In Cohn's optimization, we use a global reference set of size $N_{\text{ref}} = 2,000$ and a candidate set $N_{\text{cand}} = 100$. 

\tikzset{every picture/.style={line width=0.75pt}} 
\vspace{-1pt}


In Fig.~\ref{fig:nrmse_nlpd_result}, we illustrate that the proposed MIS method consistently achieves the lowest NRMSE and NLPD across all three SRTM datasets and fleet sizes. The proposed method reduces prediction error rapidly during early 
iterations and maintains stable performance. 
In contrast, random waypoint planning exhibits higher variability and less consistent convergence, especially in NLPD. 
Lawnmower planning provides structured coverage but generally improves more slowly and remains less sample-efficient. The SGP-based IPP~~\cite{jakkala2024multi} 
is competitive in several settings, but the proposed MIS 
reports more accurate results significantly faster. 

As the fleet size increases, 
the MIS method 
remains stable and maintains strong predictive performance, while 
baseline methods show larger fluctuations and performance degradation. This indicates that 
MIS preserves estimation quality as the team scales.

Qualitative results 
for individual agents further support these trends as shown in Fig.~\ref{fig:spatial_mean_traj}. The 
per-agent predictive mean produced by the MIS more closely matches the ground truth, while its predictive variance is lower and spatially uniform. 
Note that 
these maps are represented for a single agent within the team. 
Thus, coordinated GP modeling with pseudo-summaries favors our method for uniform uncertainty quantification in the entire field. 
In contrast, baseline methods tend to produce uneven spatial coverage 
with 
overestimated or underestimated uncertainty regions. 

Regarding computation time, the proposed method requires on average
$\{0.26,0.51,0.92\}$~\unit{sec} for distributed GP training and $\{0.48,0.95, 1.58\}$~\unit{sec} for distributed Cohn's sample selection for fleet sizes $M=\{4, 9, 16\}$ respectively with the same computational resources among different fleet sizes.
The relatively low GP update time is achieved through warm-started pxADMM optimization and pseudo-dataset exchange.  
Baseline methods perform neither distributed GP retraining nor distributed acquisition optimization at each iteration, and therefore incur lower online planning cost. 
Nevertheless, the proposed method achieves rapid gains in predictive accuracy and uncertainty reduction within only a few planning iterations, indicating that the additional computation time is justified. 

\section{Conclusion} \label{sec:Conclusion}

In this work, we formulate a Multi-Agent Informative Sampling (MIS) method that integrates distributed GP training, distributed Cohn's utility optimization, and trajectory planning. 
Numerical experiments indicate that the 
MIS demonstrates improved sample efficiency, faster reduction in prediction error, and better uncertainty calibration 
across multiple fleet sizes. 



\normalem
\bibliographystyle{IEEEtran}
\bibliography{reference}

\end{document}